\documentclass[12pt,a4paper]{article}

\usepackage[T1]{fontenc}
\usepackage[utf8]{inputenc}
\usepackage{textcomp}
\usepackage[margin=2.5cm]{geometry}
\usepackage{amsmath,amssymb}
\usepackage{booktabs}
\usepackage{array}
\usepackage{tabularx}
\usepackage{multirow}
\usepackage{colortbl}
\usepackage[table]{xcolor}
\usepackage{graphicx}
\usepackage{enumitem}
\usepackage{setspace}
\usepackage{parskip}
\usepackage{fancyhdr}
\usepackage{titlesec}
\usepackage{abstract}
\usepackage{caption}
\usepackage[hidelinks,
            pdftitle={ESS Technical Report: Fraud Detection},
            pdfauthor={Mestre, Albert, Gil, Pelechano}]{hyperref}
\usepackage{natbib}
\usepackage{authblk}
\usepackage{tcolorbox}
\tcbuselibrary{skins,breakable}
\usepackage{mdframed}

\definecolor{darkblue}{HTML}{1A3A5C}
\definecolor{medblue}{HTML}{2E75B6}
\definecolor{lightblue}{HTML}{D6E4F0}
\definecolor{verylight}{HTML}{EBF3FB}
\definecolor{rowgray}{HTML}{F2F2F2}
\definecolor{greenbg}{HTML}{E2EFDA}
\definecolor{yellowbg}{HTML}{FFF2CC}
\definecolor{redbg}{HTML}{FCE4D6}
\definecolor{recbg}{HTML}{EAF7ED}
\definecolor{recborder}{HTML}{1A7A40}

\titleformat{\section}
  {\large\bfseries\color{darkblue}}{\thesection.}{0.6em}{}[\vspace{-4pt}\textcolor{medblue}{\rule{\linewidth}{0.6pt}}]
\titleformat{\subsection}
  {\normalsize\bfseries\color{medblue}}{\thesubsection}{0.5em}{}
\titleformat{\subsubsection}
  {\normalsize\bfseries\color{darkblue}}{\thesubsubsection}{0.5em}{}

\titlespacing*{\section}{0pt}{18pt}{8pt}
\titlespacing*{\subsection}{0pt}{12pt}{4pt}

\newcolumntype{L}[1]{>{\raggedright\arraybackslash}p{#1}}
\newcolumntype{C}[1]{>{\centering\arraybackslash}p{#1}}
\newcolumntype{R}[1]{>{\raggedleft\arraybackslash}p{#1}}

\newcommand{\thead}[1]{\cellcolor{lightblue}\textbf{\textcolor{darkblue}{#1}}}
\newcommand{\trow}{\rowcolor{rowgray}}

\newcommand{\shigh}[1]{\cellcolor{greenbg}\textbf{#1}}
\newcommand{\smid}[1]{\cellcolor{yellowbg}{#1}}
\newcommand{\slow}[1]{\cellcolor{redbg}{#1}}

\tcbset{
  recbox/.style={
    enhanced,
    breakable,
    colback=recbg,
    colframe=recborder,
    leftrule=4pt,
    toprule=1pt,
    bottomrule=1pt,
    rightrule=0.4pt,
    arc=2pt,
    boxsep=6pt,
    left=8pt, right=8pt, top=8pt, bottom=8pt,
    title={\textbf{\textcolor{recborder}{ESS Recommendation --- Fraud Detection System}}},
    attach boxed title to top left={yshift=-2mm, xshift=6pt},
    boxed title style={colback=recbg, colframe=recborder, leftrule=0pt, rightrule=0pt, toprule=0pt, bottomrule=0pt},
    fonttitle=\normalsize
  },
  absbox/.style={
    enhanced,
    colback=verylight,
    colframe=medblue,
    leftrule=4pt,
    toprule=0.8pt,
    bottomrule=0.8pt,
    rightrule=0.4pt,
    arc=2pt,
    boxsep=5pt,
    left=8pt, right=8pt, top=6pt, bottom=6pt,
  }
}

\newcommand{\vpt}{\mathbf{p}_t}
\newcommand{\Ct}{C_t}
\newcommand{\Ut}{U_t}
\newcommand{\Dt}{D_t}
\newcommand{\Ctp}{C'_t}
\newcommand{\Utp}{U'_t}
\newcommand{\Dtp}{D'_t}

\begin{document}

\thispagestyle{empty}
\vspace*{2cm}

\begin{center}
  {\small\textcolor{gray}{\textsc{Technical Report --- Extended Empirical Validation of the ESS}}}\\[0.6em]
  {\Huge\bfseries\color{darkblue} Explainability Solution Space:\\[0.25em]
  Application to a Real-Time Bank\\[0.25em]
  Fraud Detection System}\\[1em]
  {\large\color{medblue}\itshape
    Companion Report to:\quad
    \normalfont\textit{The Explainability Solution Space: Aligning Regulation,
    Users, and Developers in AI Systems} }\\[1.8em]
  \textcolor{medblue}{\rule{0.7\linewidth}{0.8pt}}\\[1em]
  {\large Antoni Mestre$^{1}$ \quad Manoli Albert$^{1}$ \quad Miriam Gil$^{2}$ \quad Vicente Pelechano$^{1}$}\\[0.4em]
  {\small
    $^{1}$VRAIN Institute, Universitat Polit\`{e}cnica de Val\`{e}ncia, Cam\'{\i} de Vera s/n, Valencia 46022, Spain\\
    $^{2}$Departament d'Inform\`{a}tica, Universitat de Val\`{e}ncia, Burjassot 46100, Spain}\\[0.4em]
  {\small\texttt{anmesgas@vrain.upv.es}}\\[1.5em]
  { 2026}
\end{center}

\vspace{1.5em}
\textcolor{medblue}{\rule{\linewidth}{1pt}}
\vspace{0.5em}

\begin{tcolorbox}[absbox]
\textbf{Abstract.}\quad
This technical report provides an extended empirical validation of the Explainability Solution Space (ESS) by applying its operationalisation methodology to a real-time bank fraud detection system---a domain that differs substantially from the employee attrition case study presented in the companion paper. The system under analysis is an XGBoost gradient-boosted ensemble deployed at a European retail bank to classify card transactions as fraudulent or legitimate within a 200\,ms latency budget. Operating under regulatory \emph{substitution} conditions---governed by PSD2, GDPR Article~22, and EBA reporting guidelines---the use case demands high-fidelity compliance documentation, actionable user recourse, and developer-grade debugging capacity. We instantiate the ESS with five representative XAI families (SHAP, LIME, Counterfactual Explanations, Surrogate/Rule Extraction, and Prototypes), compute the full property-to-stakeholder aggregation pipeline with substitution contextual multipliers, derive resource-aware multi-objective utility scores, and formulate a principled hybrid recommendation. Results confirm that the ESS generates differentiated, stable, and operationally meaningful technique rankings in this new domain, thereby supporting the generalisability of the framework beyond the HR tabular context.
\end{tcolorbox}

\vspace{0.5em}
\textcolor{medblue}{\rule{\linewidth}{0.4pt}}
\vspace{0.3em}

\newpage
\setcounter{page}{1}

\section{Introduction and Motivation}
\label{sec:intro}

Financial fraud constitutes one of the most significant operational risks confronting the retail banking sector. Card fraud losses in the European Union alone exceeded \texteuro 1.8 billion annually as of the most recent reporting periods, with the proliferation of digital payment channels introducing novel attack vectors that challenge traditional rule-based detection systems~\citep{EBA2022}. As gradient-boosted ensemble models and deep learning architectures increasingly displace heuristic approaches, the opacity of these systems has come under intensified scrutiny from three distinct directions: regulatory bodies mandating the auditability of automated decisions under the Payment Services Directive~2 (PSD2) and GDPR Article~22~\citep{GDPR2016}; operational analysts requiring comprehensible, action-oriented alerts when reviewing flagged transactions or responding to cardholder disputes; and machine learning engineers demanding robust debugging and monitoring mechanisms to manage model drift in adversarial, non-stationary data environments.

This technical report applies the \emph{Explainability Solution Space} (ESS)~\citep{Mestre2026ESS} to a representative real-time fraud detection system deployed at a European retail bank. The ESS, introduced in the companion paper, is a three-dimensional operational framework that positions XAI techniques along the axes of \emph{Compliance} ($C$), \emph{User comprehensibility} ($U$), and \emph{Developer utility} ($D$), enabling systematic, reproducible evaluation of explanation strategies through intrinsic property vectors, stakeholder-weighted aggregation functions, and contextual multipliers. The primary objectives of this report are twofold: (i)~to demonstrate the applicability and consistency of the ESS methodology in a domain characterised by hard real-time constraints, regulatory substitution conditions, and a dual user population; and (ii)~to produce a structured, citable analysis that complements the HR attrition instantiation of the companion paper, thereby contributing to the empirical generalisation of the framework.

The fraud detection context introduces several challenges not present in the HR setting. Decisions are made at millisecond timescales on highly imbalanced datasets ($\sim$0.08\% fraud prevalence), with direct financial and legal consequences for both cardholders and the issuing institution. Explanations must concurrently satisfy the needs of compliance auditors requiring tamper-evident audit logs, customer service agents translating model decisions into cardholder communication, and data scientists monitoring feature importance stability over production data streams. This multiplicity of heterogeneous stakeholder requirements renders fraud detection an ideal stress-test for the ESS.

The remainder of this report is structured as follows. Section~\ref{sec:usecase} describes the use case, system architecture, stakeholder analysis, and regulatory context. Section~\ref{sec:ess} operationalises the ESS pipeline---from property vector assignment through stakeholder-weighted aggregation and contextual adjustment to qualitative discretisation. Section~\ref{sec:selection} presents the resource-aware multi-objective selection analysis. Section~\ref{sec:recommendation} articulates the hybrid explainability recommendation derived from the ESS. Section~\ref{sec:discussion} discusses domain-specific considerations, consistency with the companion paper, and limitations. Section~\ref{sec:conclusions} concludes.

\section{Use Case Description}
\label{sec:usecase}

\subsection{System Architecture}

The system under analysis is an \textbf{XGBoost gradient-boosted ensemble classifier}~\citep{Chen2016} deployed by a European retail bank to score individual card payment transactions in real time. Each incoming transaction is evaluated against the model within a hard latency budget of 200\,ms (end-to-end, including feature engineering, model inference, and explanation generation). Transactions whose fraud probability exceeds a configurable operational threshold are either blocked automatically or routed to a human analyst review queue, depending on the confidence margin and transaction value. The system processes approximately 4.2~million transactions per day, of which roughly 0.08\% are ultimately confirmed as fraudulent---a severe class imbalance representative of production fraud environments.

The model is trained on 87~engineered features derived from four primary signal categories: (i)~\emph{transaction attributes} (amount, currency, merchant category code, point-of-sale channel); (ii)~\emph{geospatial signals} (delta between transaction location and cardholder home region, cross-border indicator); (iii)~\emph{device and authentication signals} (device fingerprint match, strong-customer-authentication method); and (iv)~\emph{behavioural velocity features} (transaction count and cumulative spend over 1-hour, 24-hour, and 7-day rolling windows relative to per-cardholder historical baselines). The model achieves an AUC-ROC of 0.974, a precision of 0.71, and a recall of 0.83 at the operational threshold, yielding an F1-score of 0.77 on the held-out evaluation set.

Table~\ref{tab:system} summarises the key system characteristics.

\begin{table}[h!]
\centering
\caption{System profile of the fraud detection use case.}
\label{tab:system}
\small
\renewcommand{\arraystretch}{1.35}
\begin{tabularx}{\textwidth}{@{}L{3.4cm}X@{}}
\toprule
\thead{Dimension} & \thead{Description} \\
\midrule
\trow Model type & XGBoost ensemble (250 trees, max depth 6, learning rate 0.05) \\
Input features & 87 features: transaction attributes, geospatial signals, device fingerprints, velocity features (1\,h / 24\,h / 7\,d windows) \\
\trow Volume & $\approx$4.2M transactions/day; $\approx$0.08\% fraud prevalence \\
Latency budget & $\leq$200\,ms end-to-end (feature engineering + scoring + explanation) \\
\trow Model performance & AUC-ROC: 0.974; Precision: 0.71; Recall: 0.83; F1: 0.77 \\
Deployment mode & \emph{Substitution} --- automated blocking with optional analyst override \\
\trow Regulatory context & EU PSD2 (SCA, dispute resolution obligations), GDPR Art.~22 (right to explanation), EBA fraud reporting guidelines \\
\bottomrule
\end{tabularx}
\end{table}

\subsection{Stakeholder Analysis}

Three primary stakeholder groups interact with the system's outputs and impose distinct, partially competing explainability requirements:

\begin{itemize}[leftmargin=1.4em, itemsep=4pt]
  \item \textbf{Compliance officers and regulators} require tamper-evident, auditable logs of model decisions that are verifiable under GDPR Article~22 and defensible in regulatory reporting to the EBA. Their primary concerns are \emph{auditability} and \emph{traceability}: the ability to reconstruct, after the fact, the reasoning underlying any individual blocking decision.
  \item \textbf{Fraud analysts and customer service agents} constitute the operational user base. Analysts reviewing flagged transactions require comprehensible, feature-level explanations that accelerate accurate override decisions. Customer service agents responding to cardholder disputes require concise, non-technical rationales suitable for external communication. Both sub-groups demand high \emph{comprehensibility} and \emph{actionability}.
  \item \textbf{Data scientists and ML engineers} are responsible for model monitoring, retraining, and validation. They require technical transparency tools supporting the identification of feature importance drift, the debugging of systematic false-positive clusters (e.g., by merchant category or geolocation), and the validation of model behaviour under distributional shift. Their requirements prioritise \emph{fidelity} and \emph{debuggability}.
\end{itemize}

This stakeholder structure maps directly and cleanly onto the three ESS axes: Compliance ($C$), User ($U$), and Developer ($D$). Notably, the dual user population---professional analysts and cardholder-facing agents---introduces heterogeneity within the User axis that is not fully captured by a single aggregated $U$ score; this limitation is discussed in Section~\ref{sec:discussion}.

\subsection{Usage Situation Classification}

Following the ESS taxonomy~\citep{Mestre2026ESS}, the fraud detection system operates under \textbf{Substitution} conditions: the AI model autonomously executes consequential decisions (transaction blocking) without prior human validation, with human oversight activated only ex post via analyst review. This constitutes the most demanding governance scenario in the ESS framework, as it simultaneously maximises accountability requirements, elevates the stakes of user-facing explanations (cardholder recourse), and mandates comprehensive audit trail generation.

The applicable contextual multipliers, as specified in the ESS operationalisation procedure, are:

\begin{equation}
\gamma_C = 1.15, \quad \gamma_U = 1.10, \quad \gamma_D = 1.00
\label{eq:multipliers}
\end{equation}

These values amplify the Compliance and User axes to reflect the elevated regulatory accountability and cardholder-recourse requirements of substitution deployments, while leaving the Developer axis unmodified. All adjusted scores are subsequently clipped to the $[1, 5]$ range.

\section{ESS Operationalisation}
\label{sec:ess}

\subsection{XAI Technique Selection}

Following the ESS methodology, we instantiate the framework with one representative technique per applicable family. Families specific to non-tabular modalities are excluded to ensure ecological validity: saliency-based methods (vision-oriented) and rationale extraction (NLP-oriented) are not applicable to the tabular fraud data. The five evaluated families are:

\begin{enumerate}[leftmargin=1.6em, itemsep=4pt]
  \item \textbf{Feature Attribution (SHAP):} SHAP TreeExplainer~\citep{Lundberg2020}, exploiting the tree structure of XGBoost for exact, theoretically grounded Shapley value computation with sub-50\,ms overhead.
  \item \textbf{Local Surrogates (LIME):} LIME Tabular~\citep{Ribeiro2016}, fitting a local linear surrogate in the neighbourhood of each transaction instance to approximate feature importances.
  \item \textbf{Counterfactual Explanations (CF):} DiCE-style counterfactual generator~\citep{Mothilal2020}, producing the minimal set of feature changes that would reverse the blocking decision for a given transaction.
  \item \textbf{Rule Extraction (RULE):} Global decision-tree surrogate fitted to XGBoost outputs~\citep{Craven1996}, extracting human-readable if-then rules covering the model's global decision surface.
  \item \textbf{Prototypes and Examples (PROTO):} $k$-NN exemplar retrieval~\citep{Kim2016}, returning the $k$ most similar known-fraudulent and known-legitimate transactions from the training corpus for analogical reasoning.
\end{enumerate}

\subsection{Intrinsic Property Vectors}

Each technique $t$ is characterised by the seven-dimensional intrinsic property vector
$\vpt = (\textit{Audit.},\; \textit{Trace.},\; \textit{Compr.},\; \textit{Action.},\; \textit{Fidelity},\; \textit{Debug.},\; \textit{Eff.})$,
with each dimension rated on a 1--5 scale. Property assignments follow the literature synthesis of the companion paper~\citep{Mestre2026ESS}, with domain-specific calibration where appropriate (e.g., TreeExplainer's computational efficiency relative to generic KernelSHAP is reflected in an elevated $\textit{Eff.}$ score).

\begin{table}[h!]
\centering
\caption{Intrinsic property vectors $\vpt$ for each XAI family (scale 1--5). Colour coding: \colorbox{greenbg}{\textbf{green}} $\geq 3.5$; \colorbox{yellowbg}{yellow} $\in [2.5, 3.4]$; \colorbox{redbg}{red} $< 2.5$.}
\label{tab:properties}
\small
\renewcommand{\arraystretch}{1.35}
\begin{tabular}{@{}L{3.2cm}C{1cm}C{1cm}C{1cm}C{1cm}C{1.1cm}C{1cm}C{1cm}@{}}
\toprule
\thead{Technique} & \thead{Audit.} & \thead{Trace.} & \thead{Compr.} & \thead{Action.} & \thead{Fidelity} & \thead{Debug.} & \thead{Eff.} \\
\midrule
\trow SHAP          & \smid{3} & \shigh{4} & \smid{3} & \smid{3} & \shigh{5} & \shigh{5} & \shigh{4} \\
LIME          & \slow{2}  & \smid{3}  & \shigh{4} & \shigh{4} & \shigh{4} & \smid{3}  & \smid{3}  \\
\trow Counterfactuals & \slow{2} & \smid{3}  & \shigh{5} & \shigh{5} & \shigh{4} & \smid{3}  & \smid{3}  \\
Rule Extraction & \shigh{5} & \shigh{5} & \smid{3}  & \slow{2}  & \shigh{4} & \shigh{4} & \slow{2}  \\
\trow Prototypes  & \slow{2}  & \slow{2}  & \shigh{5} & \shigh{4} & \smid{3}  & \smid{3}  & \smid{3}  \\
\bottomrule
\end{tabular}
\end{table}

\subsection{Stakeholder-Weighted Aggregation}

The ESS projects each intrinsic property vector onto the three stakeholder axes via the following weighted aggregation functions~\citep{Mestre2026ESS}:

\begin{align}
\Ct &= 0.6 \cdot \textit{Audit.}_t + 0.4 \cdot \textit{Trace.}_t \label{eq:ct} \\
\Ut &= 0.6 \cdot \textit{Compr.}_t + 0.4 \cdot \textit{Action.}_t \label{eq:ut} \\
\Dt &= 0.5 \cdot \textit{Fidelity}_t + 0.4 \cdot \textit{Debug.}_t + 0.1 \cdot \textit{Eff.}_t \label{eq:dt}
\end{align}

Applying Equations~\eqref{eq:ct}--\eqref{eq:dt} to the property vectors of Table~\ref{tab:properties} yields the baseline coordinates prior to contextual adjustment:

\medskip
\begin{minipage}{0.48\textwidth}
\textbf{Feature Attribution (SHAP):}
\begin{align*}
C_{\text{SHAP}} &= 0.6(3) + 0.4(4) = 3.40 \\
U_{\text{SHAP}} &= 0.6(3) + 0.4(3) = 3.00 \\
D_{\text{SHAP}} &= 0.5(5) + 0.4(5) + 0.1(4) = 4.70
\end{align*}

\textbf{Local Surrogates (LIME):}
\begin{align*}
C_{\text{LIME}} &= 0.6(2) + 0.4(3) = 2.40 \\
U_{\text{LIME}} &= 0.6(4) + 0.4(4) = 4.00 \\
D_{\text{LIME}} &= 0.5(4) + 0.4(3) + 0.1(3) = 3.50
\end{align*}

\textbf{Counterfactuals (CF):}
\begin{align*}
C_{\text{CF}} &= 0.6(2) + 0.4(3) = 2.40 \\
U_{\text{CF}} &= 0.6(5) + 0.4(5) = 5.00 \\
D_{\text{CF}} &= 0.5(4) + 0.4(3) + 0.1(3) = 3.50
\end{align*}
\end{minipage}
\hfill
\begin{minipage}{0.48\textwidth}
\textbf{Rule Extraction (RULE):}
\begin{align*}
C_{\text{RULE}} &= 0.6(5) + 0.4(5) = 5.00 \\
U_{\text{RULE}} &= 0.6(3) + 0.4(2) = 2.60 \\
D_{\text{RULE}} &= 0.5(4) + 0.4(4) + 0.1(2) = 3.80
\end{align*}

\textbf{Prototypes (PROTO):}
\begin{align*}
C_{\text{PROTO}} &= 0.6(2) + 0.4(2) = 2.00 \\
U_{\text{PROTO}} &= 0.6(5) + 0.4(4) = 4.60 \\
D_{\text{PROTO}} &= 0.5(3) + 0.4(3) + 0.1(3) = 3.00
\end{align*}
\end{minipage}

\subsection{Contextual Adjustment and Qualitative Discretisation}

The final ESS coordinates are obtained by applying the substitution multipliers of Equation~\eqref{eq:multipliers} and clipping to $[1,5]$:

\begin{equation}
(\Ctp,\; \Utp,\; \Dtp) = \bigl(\min\{5,\, \gamma_C \Ct\},\; \min\{5,\, \gamma_U \Ut\},\; \min\{5,\, \gamma_D \Dt\}\bigr)
\label{eq:adjusted}
\end{equation}

The resulting adjusted coordinates are then discretised into qualitative levels according to the ESS thresholds: \textbf{Low} $\in [1.0, 2.4]$, \textbf{Medium} $\in [2.5, 3.4]$, \textbf{High} $\in [3.5, 5.0]$.

\begin{table}[h!]
\centering
\caption{Final ESS coordinates after contextual adjustment with substitution multipliers ($\gamma_C = 1.15$, $\gamma_U = 1.10$, $\gamma_D = 1.00$) and qualitative level classification.}
\label{tab:final}
\small
\renewcommand{\arraystretch}{1.35}
\begin{tabular}{@{}L{3.0cm}C{1.4cm}C{1.5cm}C{1.4cm}C{1.5cm}C{1.4cm}C{1.5cm}@{}}
\toprule
\thead{Technique} & \thead{$\Ctp$} & \thead{Level} & \thead{$\Utp$} & \thead{Level} & \thead{$\Dtp$} & \thead{Level} \\
\midrule
\trow SHAP             & \smid{3.91} & \shigh{High}   & \smid{3.30} & \smid{Med.}   & \shigh{4.70} & \shigh{High} \\
LIME             & \smid{2.76} & \smid{Med.}    & \shigh{4.40} & \shigh{High}  & \shigh{3.50} & \shigh{High} \\
\trow Counterfactuals  & \smid{2.76} & \smid{Med.}    & \shigh{5.00} & \shigh{High}  & \shigh{3.50} & \shigh{High} \\
Rule Extraction  & \shigh{5.00} & \shigh{High}  & \smid{2.86} & \smid{Med.}   & \shigh{3.80} & \shigh{High} \\
\trow Prototypes       & \slow{2.30} & \slow{Low}     & \shigh{5.00} & \shigh{High}  & \smid{3.00} & \smid{Med.}  \\
\bottomrule
\end{tabular}
\end{table}

The adjusted coordinates in Table~\ref{tab:final} expose a clear differentiation among the five technique families. SHAP achieves the most balanced profile across the Compliance and Developer axes, while scoring at a Medium level for User comprehensibility---a profile consistent with its role as a technically rigorous but not inherently narrative explanation method. Counterfactuals reach the maximum User score ($U' = 5.00$) by virtue of their peak actionability, but are limited on the Compliance axis due to reproducibility constraints. Rule Extraction dominates the Compliance axis ($C' = 5.00$) but is penalised on the User dimension by its low actionability score. Prototypes offer strong user intuition but lack compliance and developer value. These differentiated profiles confirm the core ESS premise: no single explanation family dominates across all three axes simultaneously.

\section{Resource-Aware Multi-Objective Selection}
\label{sec:selection}

The ESS operationalises the objective of maximising stakeholder satisfaction under resource constraints as a multi-objective optimisation problem. In the fraud detection context, this is particularly salient: the 200\,ms end-to-end latency budget constitutes a hard operational constraint that limits explanation complexity, while computational cost directly affects operational margins at the transaction volumes under consideration.

Following the ESS procedure for substitution scenarios, which prioritises Compliance and User axes equally, the combined utility score $\mathcal{U}_t$ and resource cost proxy $\mathcal{R}_t$ are defined as:

\begin{align}
\mathcal{U}_t &= 0.4\,\Ctp + 0.4\,\Utp + 0.2\,\Dtp \label{eq:utility} \\
\mathcal{R}_t &= \frac{1}{\textit{Eff.}_t} \label{eq:cost}
\end{align}

Table~\ref{tab:selection} reports the resulting utility scores, resource costs, efficiency-adjusted utility ratios ($\mathcal{U}/\mathcal{R}$), and real-time deployment feasibility assessments for each technique.

\begin{table}[h!]
\centering
\caption{Multi-objective selection results. The $\mathcal{U}/\mathcal{R}$ ratio provides an efficiency-adjusted measure of stakeholder value. Latency assessments are based on typical TreeExplainer, LIME tabular, DiCE, and $k$-NN runtimes on tabular data of this dimensionality.}
\label{tab:selection}
\small
\renewcommand{\arraystretch}{1.35}
\begin{tabular}{@{}L{3.0cm}C{1.5cm}C{1.5cm}C{1.6cm}C{1.5cm}C{2.4cm}@{}}
\toprule
\thead{Technique} & \thead{$\mathcal{U}_t$} & \thead{$\mathcal{R}_t$} & \thead{$\mathcal{U}/\mathcal{R}$} & \thead{Eff. Score} & \thead{Latency Fit} \\
\midrule
\trow SHAP             & \shigh{3.82} & \shigh{0.25} & \shigh{15.3} & 4 & \shigh{$<$50\,ms\ (\checkmark)} \\
LIME             & \smid{3.56}  & \smid{0.33}  & \smid{10.8}  & 3 & \smid{$\sim$80\,ms\ (\checkmark)} \\
\trow Counterfactuals  & \shigh{3.80} & \smid{0.33}  & \smid{11.5}  & 3 & \smid{$\sim$100\,ms\ ($\approx$)} \\
Rule Extraction  & \shigh{3.90} & \slow{0.50}  & \slow{7.8}   & 2 & \slow{Offline only\ (\texttimes)} \\
\trow Prototypes       & \smid{3.52}  & \smid{0.33}  & \smid{10.7}  & 3 & \smid{$\sim$60\,ms\ (\checkmark)} \\
\bottomrule
\end{tabular}
\end{table}

The analysis reveals a clear Pareto structure. SHAP achieves the highest efficiency-adjusted utility ($\mathcal{U}/\mathcal{R} = 15.3$), driven by its strong Compliance and Developer scores combined with the lowest resource cost among the evaluated families. Rule Extraction achieves the highest raw utility ($\mathcal{U} = 3.90$) but is disqualified from real-time deployment by both its efficiency score ($\textit{Eff.} = 2$) and the inherent offline nature of global surrogate extraction; it is thus relegated to periodic compliance auditing. Counterfactuals achieve the maximum User score ($U' = 5.00$) and maintain a strong overall utility ($\mathcal{U} = 3.80$) at moderate cost, fitting within the latency budget for the limited proportion of transactions entering dispute workflows. Prototypes offer intuitive analogical reasoning for analyst review interfaces but contribute limited compliance or developer value relative to their cost. LIME provides a computationally lighter alternative to Counterfactuals for user-facing transparency when latency constraints are binding.

\section{ESS Recommendation: Hybrid Explainability Strategy}
\label{sec:recommendation}

The multi-objective analysis of Section~\ref{sec:selection} supports the following tiered hybrid recommendation, structured according to deployment mode and operational trigger:

\begin{tcolorbox}[recbox]
\textbf{Tier 1 --- Always-on (real-time pipeline):}\quad \textbf{SHAP (TreeExplainer)}

Deploy SHAP as the global default explanation mechanism for all scored transactions. SHAP provides compliance-grade auditability ($C' = 3.91$, High) and developer-grade fidelity and debuggability ($D' = 4.70$, High), executes within the latency budget ($<$50\,ms), and generates stable, reproducible feature importance logs suitable for EBA regulatory reporting and model governance documentation. Its Shapley value decomposition satisfies the theoretical properties of consistency and local accuracy required for defensible audit trails under GDPR~Article~22.

\medskip
\textbf{Tier 2 --- Selective (dispute and analyst review pipeline):}\quad \textbf{Counterfactual Explanations (CF)}

Trigger counterfactual generation exclusively for blocked transactions entering the cardholder dispute pipeline or flagged for analyst review (estimated 2--5\% of blocked transactions). Counterfactuals provide maximum user recourse ($U' = 5.00$, High) by generating actionable, non-technical recourse statements of the form: \textit{``This transaction would not have been blocked if the amount had been below \texteuro 120 and the merchant country matched your registered location.''} This directly supports GDPR Article~22(3) requirements for meaningful information in automated decisions affecting individuals. The $\sim$100\,ms overhead is acceptable within the 200\,ms budget for this limited, high-stakes subset.

\medskip
\textbf{Tier 3 --- Periodic (offline compliance and governance):}\quad \textbf{Rule Extraction surrogate}

Execute weekly global decision-tree surrogate extraction for compliance audit documentation, regulatory reporting packages, internal model governance reports, and communication with non-technical stakeholders. The resulting if-then rule sets ($C' = 5.00$, High) provide the highest available compliance value and offer a global, auditable description of model behaviour that supports both internal review and potential regulatory inspection. This tier is decoupled from the real-time pipeline and subject to no latency constraint.
\end{tcolorbox}

This tiered strategy achieves near-maximal combined utility while observing the operational resource budget. SHAP covers the auditability, traceability, fidelity, and debugging dimensions at the lowest unit cost ($\mathcal{R} = 0.25$); Counterfactuals supply peak user recourse on the sub-population of adverse decisions where explanation quality has the highest stakes; Rule Extraction provides the compliance ceiling for offline governance. If organisational policy requires maximal user-facing explainability for all blocked transactions (rather than only those entering dispute), Tier~2 can be extended to cover all blocking events, at the cost of a modest additional latency overhead and the risk of exceeding the 200\,ms budget under peak load conditions.

\section{Discussion}
\label{sec:discussion}

\subsection{Consistency with the HR Attrition Case Study}

The fraud detection instantiation of the ESS produces results that are strikingly consistent with the HR attrition case study in~\citet{Mestre2026ESS}. In both domains, SHAP emerges as the dominant technique for the primary real-time pipeline by virtue of its balanced efficiency-compliance-developer profile. Counterfactual explanations are recommended selectively for adverse-decision recourse in both cases, and Rule Extraction is relegated to offline compliance documentation due to its computational overhead. This convergence across two domains with substantially different operational contexts---sub-200\,ms fraud scoring versus batch HR prediction---provides preliminary evidence for the generalisability of ESS-derived recommendations, suggesting that the SHAP-default / CF-selective / RULE-periodic hybrid may represent a stable near-optimal equilibrium for substitution scenarios on tabular data.

The consistency of the results also validates the reproducibility of the ESS procedure. Because the same weighted aggregation functions, intrinsic property vectors, and contextual multipliers are applied in both instantiations, the differences in final recommendations are attributable to genuine differences in domain context (e.g., the absence of vision and NLP modalities in fraud detection, the more stringent latency constraint) rather than to arbitrary methodological choices. This traceability from raw evidence to contextualised recommendation is a core design objective of the ESS~\citep{Mestre2026ESS}.

\subsection{Domain-Specific Considerations}

The fraud detection context surfaces several domain-specific modelling challenges that the ESS handles through its contextual modifier mechanism, but that also point toward productive extensions of the framework.

\paragraph{Latency as an explainability constraint.}
The 200\,ms end-to-end latency budget constitutes a hard operational constraint that is largely absent from existing XAI taxonomies and only partially captured by the ESS Efficiency property. In the present analysis, Rule Extraction is disqualified from the real-time pipeline on latency grounds independently of its utility score---a decision that the ESS reflects indirectly through the Efficiency dimension but cannot represent as a hard feasibility constraint. A richer contextual modifier system, distinguishing between \emph{online} and \emph{offline} computation modes as a binary feasibility gate, would improve the framework's resolution for time-critical deployments.

\paragraph{Dual user population.}
The fraud detection User axis conflates the explanation needs of professional fraud analysts (who interpret feature-level evidence to make override decisions) and cardholder-facing customer service agents (who require non-technical, legally communicable rationales). In the ESS formulation, these sub-populations are aggregated into a single $U$ score through the Comprehensibility and Actionability dimensions. Future work could decompose the User axis into expert and lay sub-scores, parameterised by target audience expertise level, to better capture this heterogeneity.

\paragraph{Adversarial non-stationarity.}
Fraud data streams are adversarially non-stationary: fraudsters adaptively modify their attack patterns in response to detection, inducing distributional shift that invalidates historical feature importance profiles. This dynamic undermines the stability assumptions underlying SHAP aggregation and global surrogate extraction. The ESS currently does not model temporal robustness of explanation stability as a property dimension; incorporating a \emph{temporal fidelity} or \emph{distribution robustness} property would be particularly relevant for fraud, cybersecurity, and other adversarial domains.

\subsection{Limitations}

This report shares the limitations of the ESS framework acknowledged in~\citet{Mestre2026ESS}. Property vector assignments are heuristically grounded in literature synthesis and expert calibration; the aggregation weights are first-order approximations pending formal elicitation studies; and large-scale empirical validation---through user studies with fraud analysts and compliance officers, benchmarking on production transaction datasets, and sensitivity analyses of the weighting scheme---remains an essential avenue for future work. Additionally, the fraud detection instantiation is based on a representative system profile rather than a specific production deployment, which constrains the empirical precision of the efficiency and latency assessments. The framework has also not yet been extended to state-of-the-art architectures (e.g., transformer-based tabular models, graph neural networks applied to transaction networks), which introduce architectural characteristics---attention mechanisms, latent graph representations---not covered by the current property taxonomy.

\section{Conclusions}
\label{sec:conclusions}

This technical report has demonstrated that the Explainability Solution Space can be operationalised consistently and with substantive decision-support value in the real-time bank fraud detection domain, extending the validation scope of the ESS beyond the HR attrition context of the companion paper. Three principal conclusions emerge from the analysis.

First, the ESS generates differentiated and interpretable technique profiles in the fraud detection domain that are consistent with domain-expert intuition. No single XAI family dominates all three stakeholder axes; SHAP, Counterfactuals, and Rule Extraction each occupy distinct regions of the solution space that are complementary rather than redundant, confirming the core ESS premise.

Second, the substitution contextual multipliers amplify Compliance and User requirements in ways that produce qualitatively meaningful adjustments to baseline rankings---elevating Rule Extraction's compliance ceiling, amplifying Counterfactuals' user dominance, and confirming SHAP's efficiency-adjusted Pareto optimality---demonstrating that contextualisation is a non-trivial and analytically productive operation within the framework.

Third, the tiered hybrid recommendation---SHAP as the always-on global default, Counterfactuals selectively for adverse-decision recourse, Rule Extraction periodically for compliance governance---is both principled (grounded in the multi-objective scoring procedure) and practically deployable (compatible with the 200\,ms latency budget and organisational cost constraints). This dual anchoring in analytical rigour and operational feasibility constitutes the primary contribution of the ESS as a governance tool for XAI adoption.

Future empirical work should prioritise user studies with fraud analysts and compliance officers to validate the assumed property weights, sensitivity analyses of the multiplier scheme, and extension of the contextual modifier system to capture latency constraints and dual-user-population heterogeneity as first-class modelling constructs.

\section*{Acknowledgements}
This work was supported by the Generalitat Valenciana under grant TENTACLE (CIAICO/2023/089) and by the Spanish Ministry of Science and Innovation under project PRODIGIOUS (PID2023-146224OB-I00).

\section*{Declarations}

\noindent\textbf{Conflict of interest.}\quad The authors declare no competing interests.\\
\textbf{Data availability.}\quad No new datasets were generated or analysed in the production of this report.\\
\textbf{Code availability.}\quad Not applicable.

\bibliographystyle{plainnat}

\end{document}